\documentclass[letterpaper, 10 pt, conference]{ieeeconf}  

\IEEEoverridecommandlockouts                              

\usepackage[citestyle=numeric-comp, bibstyle=ieee, maxbibnames=6, minbibnames=6, backend=biber]{biblatex}

\makeatletter
\let\oldlabelindent\labelindent
\let\labelindent\relax
\makeatother

\usepackage{pgfplots}
\usepackage{tikz}
\usepgfplotslibrary{groupplots,dateplot}
\usetikzlibrary{patterns,shapes.arrows,spy}
\pgfplotsset{compat=newest}
\usepgfplotslibrary{groupplots}
\usepackage{enumitem}
\usepgfplotslibrary{fillbetween}

\usepackage{xcolor}

\usepackage{booktabs}  
\usepackage{amsmath}   
\usepackage{array}     

\usepackage{wrapfig}
\usepackage{graphicx}  

\usepackage{multirow}
\usepackage{amssymb}

\usepackage[hidelinks]{hyperref}

\usepackage[inkscapelatex=false]{svg}

\usetikzlibrary{shapes, arrows.meta, positioning, bending, graphs, graphs.standard,pgfplots.fillbetween}
\usetikzlibrary{arrows,decorations.pathmorphing,positioning,fit,trees,shapes,shadows,automata,calc} 
\usetikzlibrary{patterns,arrows,arrows.meta,calc,shapes,shadows,decorations.pathmorphing,decorations.pathreplacing,automata,shapes.multipart,positioning,shapes.geometric,fit,circuits,trees,shapes.gates.logic.US,fit, matrix,arrows.meta, quotes}
\usetikzlibrary{backgrounds,scopes,spy}

\pgfplotsset{compat=newest}
\def\addlegendimage{\csname pgfplots@addlegendimage\endcsname}

\pgfdeclarelayer{background}
\pgfdeclarelayer{foreground}
\pgfsetlayers{background,main,foreground}

\definecolor{darkgray176}{RGB}{176,176,176}
\definecolor{gray}{RGB}{128,128,128}
\definecolor{lightgray204}{RGB}{204,204,204}
\definecolor{mediumturquoise62173218}{RGB}{62,173,218}
\definecolor{salmon253114111}{RGB}{253,114,111}
\definecolor{sandybrown25315671}{RGB}{253,156,71}

\definecolor{submodel1Color}{HTML}{E05F15}
\definecolor{submodel2Color}{HTML}{07742D}
\definecolor{compositeModelColor}{HTML}{4F359B}
\definecolor{trueModelColor}{HTML}{130303}

\definecolor{lightSubmodel1Color}{HTML}{F19B6A}
\definecolor{lightSubmodel2Color}{HTML}{0cd452}
\definecolor{lightCompositeModelColor}{HTML}{a494c4}

\definecolor{VLALight}{HTML}{96BBD9}
\definecolor{VLADark}{HTML}{4180B4}
\definecolor{VLAPSLight}{HTML}{F19955}
\definecolor{VLAPSDark}{HTML}{ED771D}

\definecolor{highFidelityLight}{HTML}{F0A8B1}
\definecolor{highFidelityDark}{HTML}{D1233B}

\definecolor{ours}{HTML}{ED771D}
\definecolor{oursTwo}{HTML}{74B3CE}
\definecolor{oursThree}{HTML}{079C72}
\definecolor{oursFour}{HTML}{6D7A88}

\definecolor{base}{HTML}{FE5655}
\definecolor{baseTwo}{HTML}{8F5F95}
\definecolor{baseThree}{HTML}{1F487E}
\definecolor{baseFour}{HTML}{6D7A88}

\definecolor{outlineGray}{gray}{0.7}
\definecolor{arrowGray}{gray}{0.4}

\definecolor{success}{HTML}{00B050}
\definecolor{failure}{HTML}{FF0000}

\usepackage[acronyms, shortcuts]{glossaries}
\glsdisablehyper

\newacronym{drl}{DRL}{deep reinforcement learning}
\newacronym{il}{IL}{imitation learning}
\newacronym{rl}{RL}{reinforcement learning}
\newacronym{mdp}{MDP}{Markov decision process}
\newacronym{pomdp}{POMDP}{partially observable Markov decision process}
\newacronym{vla}{VLA}{vision-language-action}
\newacronym{vlaps}{VLAPS}{vision-language-action planning and search}
\newacronym{llm}{LLM}{large language model}
\newacronym{vlm}{VLM}{vision language model}
\newacronym{mcts}{MCTS}{Monte Carlo tree search}
\newacronym{mpc}{MPC}{model-predictive control} 

\makeatletter
\let\labelindent\oldlabelindent
\makeatother

\newcommand{\pomdp}{\mathcal{M}}
\newcommand{\mdp}{\mathcal{M}}
\newcommand{\state}{s}
\newcommand{\stateSet}{S}
\newcommand{\action}{a}
\newcommand{\actionSet}{A}
\newcommand{\transitionFunction}{T}
\newcommand{\rewardFunction}{R}

\newcommand{\worldModel}{\widehat{\mathcal{M}}}
\newcommand{\worldModelState}{\widehat{\state}}

\newcommand{\task}{\mathcal{T}}
\newcommand{\languageInstruction}{L_{\task}}
\newcommand{\finalStateSet}{\stateSet_{\task}}

\newcommand{\actionDimension}{n}

\newcommand{\image}{I}

\newcommand{\macroAction}{u}
\newcommand{\macroActionSet}{U}
\newcommand{\macroActionLibrary}{\Phi}

\newcommand{\macroActionHorizon}{H}
\newcommand{\treeNode}{v}

\newcommand{\policy}{\pi}

\newcommand{\vlaPrior}{\psi_{\macroActionLibrary_{\treeNode}}}
\newcommand{\vlaDistribution}{P_{vla}}
\newcommand{\samplingDistributionTemperature}{\alpha}
\newcommand{\distanceMeasure}{\rho}
\newcommand{\samplingDistribution}{\beta_{\macroActionLibrary}}

\newcommand{\numSamples}{k}

\newcommand{\numVisits}{N}

\makeatletter
\newcommand{\shiftedfootnote}[2][-2mm]{%
  \begingroup
  \renewcommand{\@makefntext}[1]{%
    \setlength{\parindent}{0pt}%
    \hspace*{#1}
    \noindent\makebox[1.8em][r]{\@makefnmark\,}##1%
  }%
  \footnote{#2}%
  \endgroup
}
\makeatother

\title{\LARGE \bf
Improving Pre-Trained Vision-Language-Action \\ Policies with Model-Based Search
}

\author{
  Cyrus Neary\textsuperscript{1,2,3}, \; Omar G. Younis\textsuperscript{1}, \; Artur Kuramshin\textsuperscript{1,2}, \; \"Ozg\"ur Aslan\textsuperscript{1,2}, \;  Glen Berseth\textsuperscript{1,2}\\
  \small \textsuperscript{1}Mila --- Quebec AI Institute, Canada,  \textsuperscript{2}Université de Montréal, Canada\\
  \small \textsuperscript{3}The University of British Columbia, Canada\\
  \small cyrus.neary@ubc.ca, \{omar.younis, artur.kuramshin, ozgur.aslan, glen.berseth\}@mila.quebec
}

\begin{document}

\maketitle
\par

\thispagestyle{empty}
\pagestyle{empty}

\begin{abstract}
    Pre-trained vision-language-action (VLA) models offer a promising foundation for generalist robot policies, but often produce brittle behaviors or unsafe failures when deployed zero-shot in out-of-distribution scenarios.
    We present \textit{Vision-Language-Action Planning \& Search (VLAPS)}---a novel framework and accompanying algorithms that embed model-based search into the inference procedure of pre-trained VLA policies to improve their performance on robotic tasks.
    Specifically, our method biases a modified Monte Carlo Tree Search (MCTS) algorithm---run using a model of the target environment---using action priors defined by the VLA policy.
    By using VLA-derived abstractions and priors in model-based search, VLAPS efficiently explores language-conditioned robotics tasks whose search spaces would otherwise be intractably large.
    Conversely, by integrating model‑based search with the VLA policy’s inference procedure, VLAPS yields behaviors that are more performant than those obtained by directly following the VLA policy’s action predictions.
    VLAPS offers a principled framework to: i) control test-time compute in VLA models, ii) leverage a priori knowledge of the robotic environment, and iii) integrate established planning and reinforcement learning techniques into the VLA inference process.
    Across all experiments, VLAPS significantly outperforms VLA-only baselines on language-specified tasks that would otherwise be intractable for uninformed search algorithms, increasing success rates by as much as \(67\) percentage points.     
\end{abstract}

\begin{keywords}%
  Generalist robot policies, vision-language-action models, inference-time compute, model-based search, Monte Carlo tree search, robot learning%
\end{keywords}

\section{Introduction}
\label{sec:intro}

Large \ac{vla} models---pre-trained on internet-scale vision and language data, as well as on diverse datasets of robot demonstrations---offer promise as a key building block for generalist robot policies that autonomously solve complex, open-ended tasks.
Their natural‐language interface enables intuitive task specification, and their multimodal reasoning capabilities support flexible and rapid adaptation to new tasks and environments.
However, current \acp{vla} are unable to reason over the potential consequences of their actions, 
and instead rely purely on imitating observed behaviors from their training data.
Myopically following \ac{vla} action outputs can thus result in brittle behaviors and unsafe failures, impeding the adoption of such models in many robotics applications.

On the other hand, model-based planning algorithms explicitly reason over future outcomes, but typically rely on carefully handcrafted heuristics to address the combinatorial explosion of the search space.
Indeed, many robotics tasks are naturally characterized by sparse rewards in large state-action spaces, making direct search without a well-designed heuristic infeasible.
It remains challenging to design heuristics that generalize across diverse robotic tasks---particularly those specified in natural language and deployed in complex, cluttered environments.

\begin{figure*}
    \centering
    \includegraphics[]{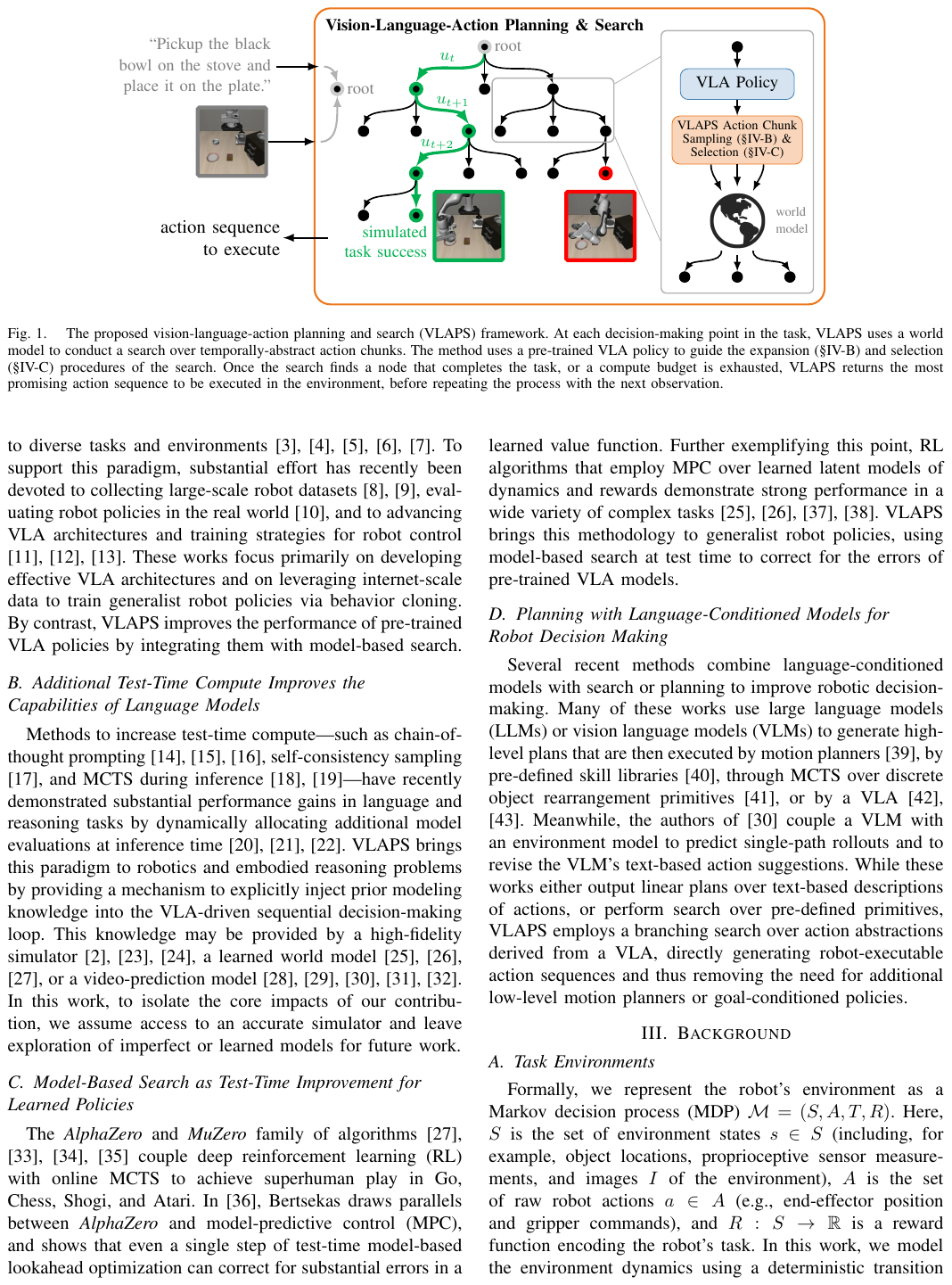}
    \caption{
    The proposed \acf{vlaps} framework. 
    At each decision-making point in the task, \ac{vlaps} uses a world model to conduct a search over temporally-abstract action chunks. 
    The method uses a pre-trained VLA policy to guide the expansion (\S \ref{sec:defining_search_space}) and selection (\S \ref{sec:biasing_selection}) procedures of the search. 
    Once the search finds a node that completes the task, or a compute budget is exhausted, \ac{vlaps} returns the most promising action sequence to be executed in the environment, before repeating the process with the next observation.
    }
    \label{fig:intro_figure}
\end{figure*}

Towards developing generalist robot policies that are flexible, robust, and capable of reasoning over long horizons, we introduce \textit{\acf{vlaps}}.
\ac{vlaps} integrates model-based search with a pre-trained \ac{vla} policy, to leverage the core strengths of both approaches to problem solving.
Figure \ref{fig:intro_figure} illustrates the approach:
\ac{vlaps} uses a pre-trained VLA to bias a \ac{mcts}-inspired algorithm \cite{swiechowski2023monte} through several distinct mechanisms.
It begins by using the \ac{vla} policy to sample contextually-relevant collections of temporally-abstract \textit{action chunks} 
at every node of the search tree, automatically refining the large action spaces typical of robotics problems into tractable, task-relevant spaces for search.
It then leverages VLA-derived priors to bias node selection during tree traversal, further focusing the search on promising behaviors suggested by the \ac{vla} policy. 
By doing so, \ac{vlaps} efficiently explores language-specified robotics tasks whose search spaces would otherwise be intractably large.
Conversely, by integrating model‑based search with the \ac{vla} model’s inference procedure, \ac{vlaps} yields policies that are more performant than those obtained by directly following the \ac{vla}’s action predictions.

We demonstrate the benefits of the proposed approach in \textit{LIBERO} \cite{liuLibero}---a simulated suite of language-specified robotic manipulation tasks.
By integrating a \ac{vla} policy with \ac{mcts}-style search, \ac{vlaps} outperforms the underlying \ac{vla}, with task success rates improving by up to \(67\) percentage points.
We also find that \ac{vlaps}' performance increases alongside that of the underlying \ac{vla}, and that it adaptively allocates more search to difficult instances.
Moreover, it can boost the performance of relatively small \ac{vla}s to levels that match much larger, sate-of-the-art models.

\section{Related Work}
\label{sec:related_work}

\subsection{\ac{vla} Models as Generalist Robot Policies}
Recent work has demonstrated that large, multimodal transformers can map natural language instructions and visual observations directly to robot actions, enabling the development of generalist robot policies that flexibly adapt to diverse tasks and environments \cite{brohan2022rt,brohan2023rt,shridhar2023perceiver,driess23a,team2024octo}.
To support this paradigm, substantial effort has recently been devoted to collecting large-scale robot datasets \cite{khazatsky2024droid,o2024open}, evaluating robot policies in the real world \cite{atreya2025roboarena}, and to advancing \ac{vla} architectures and training strategies for robot control \cite{kim2024openvla,black2024pi_0,pertsch2025fast}.
These works focus primarily on developing effective \ac{vla} architectures and on leveraging internet-scale data to train generalist robot policies via behavior cloning.
By contrast, \ac{vlaps} improves the performance of pre-trained \ac{vla} policies by integrating them with model-based search.

\subsection{Additional Test-Time Compute Improves the \\ Capabilities of Language Models}
Methods to increase test-time compute---such as chain-of-thought prompting \cite{weiCainOfThought,yao2023tree,xiang2025towards}, self-consistency sampling \cite{wang2023selfconsistency}, and \ac{mcts} during inference \cite{hao2023reasoning,feng2023alphazerolike}---have recently demonstrated substantial performance gains in language and reasoning tasks by dynamically allocating additional model evaluations at inference time \cite{guo2025deepseek,ji2025test,li2025system}.
\ac{vlaps} brings this paradigm to robotics and embodied reasoning problems by providing a mechanism to explicitly inject prior modeling knowledge into the \ac{vla}-driven sequential decision-making loop. 
This knowledge may be provided by a high-fidelity simulator \cite{makoviychuk2021isaac,liuLibero,zhu2020robosuite}, a learned world model \cite{Hafner2020Dream,hafner2021mastering,schrittwieser2020mastering}, or a video-prediction model \cite{yang2024learning,bruce2024genie,feng2025reflective,parkerholder2024genie2,du2024video}.
In this work, to isolate the core impacts of our contribution, we assume access to an accurate simulator and leave exploration of imperfect or learned models for future work.

\subsection{Model-Based Search as Test-Time Improvement for \\ Learned Policies}
The \textit{AlphaZero} and \textit{MuZero} family of algorithms \cite{silver2016mastering,silver2017mastering,silver2018general,schrittwieser2020mastering} couple deep \ac{rl} with online \ac{mcts} to achieve superhuman play in Go, Chess, Shogi, and Atari.
In \cite{bertsekas}, Bertsekas draws parallels between \textit{AlphaZero} and \ac{mpc}, and shows that even a single step of test-time model-based lookahead optimization can correct for substantial errors in a learned value function.
Further exemplifying this point, \ac{rl} algorithms that employ \ac{mpc} over learned latent models of dynamics and rewards demonstrate strong performance in a wide variety of complex tasks  \cite{Hafner2020Dream,hafner2021mastering,hansen22a,hansen2024tdmpc}.
\ac{vlaps} brings this methodology to generalist robot policies, using model-based search at test time to correct for the errors of pre-trained \ac{vla} models.

\subsection{Planning with Language-Conditioned Models for \\ Robot Decision Making}
Several recent methods combine language-conditioned models with search or planning to improve robotic decision-making. 
Many of these works use \acp{llm} or \acp{vlm} to generate high-level plans that are then executed by motion planners \cite{dalal2023planseqlearn}, by pre-defined skill libraries \cite{ahn2022can}, through \ac{mcts} over discrete object rearrangement primitives \cite{changLGMCTS}, or by a \ac{vla} \cite{shi2025hi,zawalski2024robotic}.
Meanwhile, the authors of \cite{feng2025reflective} couple a \ac{vlm} with an environment model to predict single-path rollouts and to revise the \ac{vlm}'s text-based action suggestions.
While these works either output linear plans over text-based descriptions of actions, or perform search over pre-defined primitives, \ac{vlaps} employs a branching search over action abstractions derived from a \ac{vla}, directly generating robot-executable action sequences and thus removing the need for additional low-level motion planners or goal-conditioned policies.
\section{Background}
\label{sec:background}

\subsection{Task Environments}
Formally, we represent the robot's environment as a \ac{mdp} \(\mdp = (\stateSet, \actionSet, \transitionFunction, \rewardFunction)\).
Here, \(\stateSet\) is the set of environment states \(\state \in \stateSet\) (including, for example, object locations, proprioceptive sensor measurements, and images \(\image\) of the environment), \(\actionSet\) is the set of raw robot actions \(\action \in \actionSet\) (e.g., end-effector position and gripper commands), and \(\rewardFunction : \stateSet \to \mathbb{R}\) is a reward function encoding the robot's task.
In this work, we model the environment dynamics using a deterministic transition function \(\transitionFunction : \stateSet \times \actionSet \to \stateSet\) which maps a state \(\state_{t} \in \stateSet\) and action \(\action_{t} \in \actionSet\) at time \(t\) to the corresponding state at the next timestep \(\state_{t+1} \in \stateSet\).

\subsection{Robotic Tasks}
We formalize the robot's task as a tuple \(\task = (\languageInstruction, \finalStateSet)\), where \(\languageInstruction\) is a natural-language instruction such as \textit{``Place the orange juice in the basket"} and \(\finalStateSet \subseteq \stateSet\) is the set of all environment states in which this task has been completed.
The sparse reward function \(\rewardFunction_{\task}\) corresponding to a particular task \(\task\) is thus defined to return \(\rewardFunction_{\task}(\state) = 1\) when \(\state \in \finalStateSet\), and \(\rewardFunction_{\task}(\state) = 0\) otherwise.
Our objective in this work is to create a policy that, given information on the current state \(s\) and a natural language instructions \(\languageInstruction\) for an arbitrary task \(\task\), outputs a sequence of actions that efficiently leads to a goal state \(\state \in \finalStateSet\).

\subsection{Vision-Language-Action Models}
A promising approach for designing such policies is to leverage \ac{vla} models---large-scale transformer networks trained to directly predict action sequences from multimodal observation histories and natural-language task instructions.
Such models are often trained by fine-tuning a pre-trained \ac{vlm} to either autoregressively predict actions or to guide a diffusion process for sampling them.
In order to learn efficiently, \ac{vla} models often discretize the continuous action space into a sequence of discrete tokens, a process known as action tokenization. 
This allows robot actions to be represented in an identical format to language tokens, thus enabling the application of architectures and training techniques originally developed for text modeling.

\subsection{Monte Carlo Tree Search}
\ac{mcts} is an algorithm for sequential decision-making that incrementally builds a search tree by simulating possible future trajectories, while automatically balancing exploration of uncertain actions with exploitation of actions whose estimated values are high \cite{swiechowski2023monte}.
MCTS is well-suited for problems with large action spaces and sparse rewards, as it focuses computational effort towards the most promising and underexplored regions of the search space.
MCTS proceeds through four main phases at each decision point:
\begin{enumerate} [itemsep=0.3em, leftmargin=0em, label={}]
    \item \textit{Selection: } Starting from the root node \(\treeNode_{0}\), recursively select child nodes according to a selection policy (e.g., PUCT \cite{silver2017mastering}) until reaching a leaf node.
    \item \textit{Expansion: } If the selected leaf node is not terminal, expand the tree by adding one or more child nodes corresponding to the leaf node's available actions.
    \item \textit{Simulation: } From each newly added child node, use a \textit{rollout policy} \(\policy_{rollout}\) to simulate a trajectory and sample Monte Carlo estimates of the expected outcome.
    \item \textit{Backpropagation: } Update the value estimates along the path from the expanded node to the root node based on the simulation results, which will in turn be used during the next selection phase.
\end{enumerate}

\section{Vision-Language-Action Planning \& Search}
\label{sec:method}

We now present \ac{vlaps}, which integrates \ac{mcts}-inspired search with a pre-trained \ac{vla} policy.
Given access to a pre-trained \ac{vla} policy and a \textit{world model} \(\worldModel\) that may be used to simulate the true operating environment \(\pomdp\), \ac{vlaps} proceeds as follows.

\subsection{An Overview of the VLAPS Search Procedure.}
Figure \ref{fig:intro_figure} illustrates an overview of the approach, while Figure \ref{fig:action_expansion} details the node expansion and action selection procedures.
At each decision point \(t\), \ac{vlaps} begins by instantiating a search tree with a root node \(\treeNode_{0}\) containing information on the current state \(\state_{t} \in \stateSet\) from the true environment \(\pomdp\).
It then searches the tree from this root node using a \ac{vla}-guided variant of the PUCT selection criterion (\S \ref{sec:biasing_selection}), until it reaches a leaf node \(\treeNode\) corresponding to a potential future state that has not yet been expanded. 
Once such a leaf node is reached, the method expands the tree by sampling a collection of action sequences from a prior distribution defined by the \ac{vla} policy (\S \ref{sec:defining_search_space}), and simulating them within the world model \(\worldModel\) to generate corresponding child nodes \(\treeNode'\). 
From each such expanded child node, \ac{vlaps} then uses the model \(\worldModel\) to simulate rollouts
of the \ac{vla} policy until either the task is complete or a maximum horizon length is reached. 
\ac{vlaps} iteratively repeats this process to expand the tree until either a goal state \(\worldModelState \in \finalStateSet\) is reached, or until a maximum number of \ac{mcts} steps has been exhausted.
In the former case, the algorithm returns the sequence of actions that reach the encountered goal state.
Otherwise, it returns the most frequently-selected action sequence from the root node, and repeats the search process from the resulting next state.

We emphasize that our approach leverages the \ac{vla} policy to address several challenges faced by search algorithms in our language-conditioned robotics task setting:

\begin{figure}[b]
    \centering
    \includegraphics[]{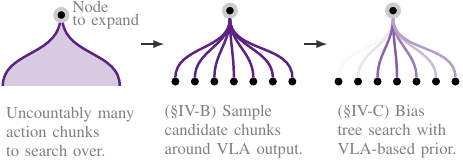}
    \caption{
        \ac{vla}-informed node expansion and action selection.
        Whenever a new node is encountered, \ac{vlaps} uses the \ac{vla} policy to sample a discrete set of candidate action chunks to search over (\ref{sec:defining_search_space}).
        Throughout the remaining search, \ac{vlaps} biases action chunk selection towards chunks similar to those output by the \ac{vla} (\ref{sec:biasing_selection}).
    }
    \label{fig:action_expansion}
\end{figure}

\begin{itemize}[itemsep=0.3em, leftmargin=0em, label={}]

    \item \textit{Refining the search space (\S \ref{sec:defining_search_space}).} 
    Directly searching over all possible robot actions is infeasible, even for algorithms like \ac{mcts} that are specifically designed to explore large search spaces.
    We accordingly present a method to refine the search space through \ac{vla}-directed sampling.
    
    \item \textit{Biasing the \ac{mcts} selection phase (\S \ref{sec:biasing_selection}).}
    Even while operating within this simplified search space, uniform random search may be computationally expensive.
    We thus introduce a method to bias the search towards actions favored by the \ac{vla} policy.
    
\end{itemize}
\vspace{-2mm}

\subsection{Automatically Defining Task-Oriented, \\ Tractable Search Spaces}
\label{sec:defining_search_space}

The state and action spaces of robotics environments are typically high-dimensional and continuous, rendering the direct application of \ac{mcts}---which is designed to sample from a finite set of possible transitions between tree nodes---challenging.
Furthermore, the size of the search tree grows exponentially in its depth, which poses a challenge for long-horizon tasks, particularly when considering decision-making timesteps that align with the short timescales of the robot's primitive actions \(\action \in \actionSet\).
To address these challenges, existing methods perform search over pre-defined subtasks or over symbolically-defined objectives~\cite{changLGMCTS}, as opposed to over individual actions.
However, such subtasks must typically be hand-designed, limiting their adaptation to new problems.

Towards addressing these challenges, \ac{vlaps} performs its search over sets of temporally-abstract \textit{action chunks} \cite{zhao2023learning} \(\macroAction_{t} = (\action_{t}, \ldots, \action_{t + \macroActionHorizon}) \in \macroActionSet\) that are flexibly defined at runtime by the outputs of the \ac{vla} policy.
By leveraging the \ac{vla} policy to generate candidate action chunks at each state of the tree search, \ac{vlaps} automatically constructs a task-relevant, state-dependent, and tractable search space.
This design also enables user control over the chunk horizon \(\macroActionHorizon\), which governs the frequency at which the \ac{vla} must be queried during search.

\subsubsection{Defining a Finite Library of Candidate Action Chunks} 
We identify the space of primitive actions \(\actionSet\) with a subset of \(\mathbb{R}^{\actionDimension}\), and the space of action chunks \(\macroActionSet\) with a subset of \(\mathbb{R}^{\macroActionHorizon \times \actionDimension}\), where \(\actionDimension\) is the dimension of the action space and \(\macroActionHorizon\) is the length of each chunk.
Relevant action chunks may, for example, consist of a short trajectory that executes a coherent robot motion.
Rather than conducting search over the entirety of \(\mathbb{R}^{\macroActionHorizon \times \actionDimension}\), \ac{vlaps} restricts planning to a finite set \(\macroActionLibrary \subseteq \macroActionSet\) of candidate action chunks.
Although this library is finite, it must be sufficiently large and expressive to support solutions across a diverse range of robotic tasks.
In our experiments, we construct \(\macroActionLibrary\) by sampling thousands of action chunks from a pre-trained \ac{vla} across a suite of tasks.
Alternatively, one could define \(\macroActionLibrary\) as the collection of all action sequences of length \(\macroActionHorizon\) appearing in the large-scale demonstration datasets used to train the \ac{vla} itself.
By constructing the candidate library from demonstrations, we restrict the search to behaviorally-relevant regions of the action space, which we hypothesize may improve the feasibility and reliability of the selected actions when deployed on physical robots.

\subsubsection{Sampling Contextually-Relevant Action Chunk Subsets}
Although the library \(\macroActionLibrary\) of action chunks is finite, it is still too large to search exhaustively from each tree node during planning.
Indeed, many candidate action chunks will likely be irrelevant at any given decision point during a task (e.g., closing the gripper will not be a useful action chunk to consider when the robot's end effector is several feet away from the object it is attempting to grasp).
To further refine the search space, we thus use the \ac{vla} policy to define a sampling distribution \(\samplingDistribution(\cdot | \image_t, \languageInstruction)\) over action chunks \(\macroAction \in \macroActionLibrary\), as in \eqref{eq:sampling_distribution} below.
Recall that \(\image_t\) denotes an image observation of the environment and \(\languageInstruction\) denotes a natural language task instruction.

To expand a leaf node \(\treeNode\) during search, \ac{vlaps} draws \(\numSamples\) samples from \(\samplingDistribution\) to propose a collection \(\macroActionLibrary_{\treeNode} \triangleq \{\macroAction^{i} | \macroAction^{i} \sim \samplingDistribution(\cdot|\image_t, \languageInstruction)\}_{i=1}^{k}\) of candidate action chunks.
By reducing the branching factor from \(|\macroActionLibrary|\) to \(\numSamples\), these action chunks constrain the local action space at the current node \(\treeNode\) and serve as the sole candidates considered for tree expansion.
Once sampled, they remain fixed for node \(\treeNode\) for the duration of the planning episode and are reused throughout the search process until a new episode begins and a fresh tree is constructed.

The distribution \(\samplingDistribution(\cdot | \image_{t}, \languageInstruction)\) uses the \ac{vla} model to bias sampling towards contextually-relevant action chunks.
As a result, action chunks that are well suited to the current stage of the task are assigned higher probability, enabling a targeted and situation-aware exploration of the action space.
More specifically, given an image observation \(\image_t\) and a language instruction \(\languageInstruction\), let \(\macroAction^{vla}(\image_t, \languageInstruction) \sim \vlaDistribution(\image_t, \languageInstruction)\) be the action chunk sampled from the \ac{vla} model.
We define the sampling distribution \(\samplingDistribution(\cdot | \image_t, \languageInstruction)\) for every \(\macroAction^{i} \in \macroActionLibrary\) to be centered around \(\macroAction^{vla}\) as 
\begin{equation}
    \samplingDistribution(\macroAction^{i} | \image_t, \languageInstruction) = (1-\epsilon)\frac{\exp \bigl(-\samplingDistributionTemperature * \distanceMeasure(\macroAction^{i}, \macroAction^{vla}) \bigr)}{\sum_{j = 1}^{|\macroActionLibrary|} \exp \bigl(-\samplingDistributionTemperature * \distanceMeasure(\macroAction^{j}, \macroAction^{vla}) \bigr) } + \frac{\epsilon}{|\macroActionLibrary|}.
    \label{eq:sampling_distribution}
\end{equation}
Here, \(\distanceMeasure(\cdot , \cdot)\) is a distance metric between action chunks; 
\(\samplingDistributionTemperature \in \mathbb{R}_{+}\) is an inverse temperature parameter that controls how sharply the distribution favors action chunks similar to the \ac{vla}'s output;
and \(\epsilon \in [0,1]\) governs epsilon-uniform exploration, ensuring that all elements \(\macroAction^{i} \in \macroActionLibrary\) are assigned nonzero probability.
While many choices of distance metric \(\distanceMeasure\) are possible, we use the Euclidean distance between flattened action chunks: \(\distanceMeasure(\macroAction^{1}, \macroAction^{2}) \triangleq ||\macroAction^{1}_{flat} - \macroAction^{2}_{flat}||_{2}\), 
where each action chunk is reshaped into a vector by flattening along its temporal dimension.

We note that this form of sampling-based action space reduction is closely related to the framework proposed in \cite{hubert2021learning}, which formally studies the impact of action subset sampling on policy improvement and evaluation algorithms in large and complex action spaces.

\subsection{Guiding Tree Traversal with a \ac{vla}-Based Prior Policy}
\label{sec:biasing_selection}

Having used the VLA policy to define \(\samplingDistribution\), and subsequently to define each node’s set \(\macroActionLibrary_{\treeNode}\) of action chunks during \textit{expansion}, we now turn to the \textit{selection} phase of the search.
In particular, \ac{vlaps} further uses the \ac{vla} policy to define a prior distribution \(\vlaPrior\) that influences the search's exploration, biasing the tree traversal toward promising subtrees preferred by the \ac{vla} policy.
Intuitively, while \S\ref{sec:defining_search_space} uses the VLA policy to constrain each node’s action chunk set \(\macroActionLibrary_{\treeNode}\) during node expansion, here we use it to bias exploration among those previously-pruned sets of action chunks.

Specifically, we use a policy-guided tree search that omits value estimates, i.e., PUCT \cite{silver2017mastering} with \(Q\equiv0\).
At each selection step of tree traversal, \ac{vlaps} chooses the next node \(\treeNode'\) by maximizing the score \eqref{eq:puct} over the node-specific candidate action chunks \(\macroAction^{i} \in \macroActionLibrary_{\treeNode}\) sampled from \(\samplingDistribution\).
\begin{equation}
     \textrm{SCORE}(\treeNode, \macroAction^{i}) = \vlaPrior(\macroAction^{i}|\image_t, \languageInstruction) \frac{\sqrt{\numVisits(\treeNode, \macroAction^{i})}}{1 + \numVisits(\treeNode, \macroAction^{i})}
    \label{eq:puct}
\end{equation}
The resulting selection rule allocates visits approximately in proportion to a prior distribution \(\vlaPrior(\macroAction^{i} | \image_t, \languageInstruction)\) derived from \ac{vla} samples, while ensuring coverage via a node-action visitation count term \(\numVisits(\treeNode, \macroAction^{i})\). 
We define \(\vlaPrior(\macroAction^{i} | \image_t, \languageInstruction)\) as a softmax-style distribution centered on the output of the \ac{vla} model, similar to the construction of \(\samplingDistribution\) in \eqref{eq:sampling_distribution}.

We note that value function estimates---which might be obtained from rollout returns as in classical \ac{mcts} \cite{swiechowski2023monte}, observation-conditioned proxies from VLMs \cite{ma2024vision}, or hybrids that combine learned critics and rollouts \cite{silver2016mastering}---could additionally be included in \eqref{eq:puct} to bias search towards promising branches.
However, learning accurate value function estimates for generalist, multi-task robot policies remains an open area of research.
In this work we choose to omit such value estimates to isolates the effect of our \ac{vla}-based prior on exploration, and to avoid reliance on potentially erroneous critics; we terminate the search upon encountering a simulated success. 
We find this variant of tree traversal to be well-suited to the studied sparse-reward setting with strong \ac{vla}-based priors.

\section{Experimental Results}
\label{sec:experiments}

To demonstrate the key benefits of \ac{vlaps}, we compare it to pre-trained \ac{vla} models on simulated vision-and-language-conditioned robotic manipulation tasks from \textit{Libero} \cite{liuLibero}.
Project code, datasets, and \ac{vla} model checkpoints are publicly available
\shiftedfootnote[-4mm]{
\begin{minipage}[t]{0.99\textwidth}
    Code: \href{https://github.com/cyrusneary/vlaps}{github.com/cyrusneary/vlaps}.\\
    Checkpoints: \href{https://huggingface.co/cyrusneary/vlaps_v1_octo_libero_finetuned}{huggingface.co/cyrusneary/vlaps\_v1\_octo\_libero\_finetuned}.\\
    Datasets: \href{https://huggingface.co/datasets/real-lab/libero_filtered_noops_rlds_dataset}{huggingface.co/datasets/real-lab/libero\_filtered\_noops\_rlds\_dataset}.
\end{minipage}
}.

\subsection{Experimental Procedures}
\label{sec:results_experimental_procedure}

\subsubsection{Evaluation Tasks and the Baseline \ac{vla} Model}
We evaluate our method on tasks drawn from all five of the Libero task suites---\textit{Libero-Spatial}, \textit{Libero-Goal}, \textit{Libero-Object}, \textit{Libero-90}, and \textit{Libero-10}---and we compare \ac{vlaps}' performance against a baseline \ac{vla}-only policy.
We note that this \ac{vla}-only baseline reflects the typical deployment of \ac{vla} models in the literature.

For both \ac{vlaps} and the \ac{vla}-only baseline policy, we use \textit{Octo} \cite{octo_2023} as the underlying \ac{vla} model.
More specifically, we finetune \textit{Octo-base-1.5} on the Libero dataset \cite{liuLibero} to predict end-effector pose and gripper commands from language instructions, a \(256\)x\(256\) fixed camera image, and a \(128\)x\(128\) wrist camera image.
The finetuning dataset comprises demonstrations from each of the Libero task suites, and we process the demonstrations to filter out all ``no-op" actions, similarly to as in \cite{kim2024openvla}.

\begin{figure}
    \centering
    \includegraphics[]{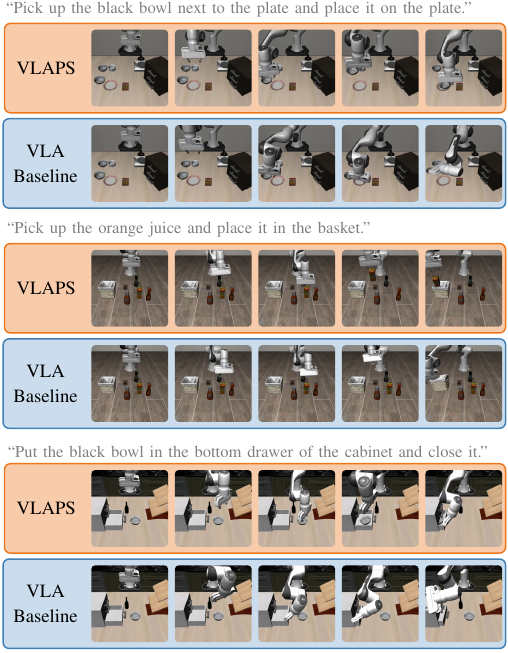}
    \vspace{-5mm}
    \caption{
        Illustrative examples where \ac{vlaps} succeeds while the underlying \ac{vla} policy fails.
        The \ac{vla} policy is \textit{Octo} finetuned for \(50\)k steps on Libero demonstration data.
        In the top two tasks, the \ac{vla} policy fails to grasp the target object, which prevents successful task completion, whereas \ac{vlaps} successfully completes the pick-and-place.
        In the bottom task, the \ac{vla} policy prematurely closes the drawer before retrieving the bowl, while \ac{vlaps} correctly places the bowl inside before closing it.
    }
    \label{fig:filmstrips}
\end{figure}

\subsubsection{\ac{vlaps} Implementation}
All experiments were conducted on an Nvidia A100 GPU. 
We used \(300\) Monte Carlo samples per search iteration, expanded \(k=10\) children per node, and allowed simulated rollouts of up to \(300\) steps. 
Action chunks had a horizon of \(H=4\), consistent with the Octo training setup, and the maximum tree search depth was set to \(100\). 
The candidate library of action chunks \(\macroActionLibrary\) was constructed by aggregating action sequences from successful runs under the \ac{vla}-only policy, normalizing these collections of action chunks per dimension, and applying K-Medoids clustering to extract a set of \(2000\) representative action chunk prototypes.
To define the action chunk distributions \(\beta_{\Phi}(\cdot)\) and \(\psi_{\Phi_{v}}(\cdot)\), we used a softmax distribution with inverse temperature parameters \(\alpha_{\beta_{\Phi}} = 10.0\) for action-chunk sampling and \(\alpha_{\psi_{\Phi_{v}}} = 5.0\) for the VLA-defined prior policy, with uniform exploration probabilities \(\epsilon_{\Phi_{\beta}} = \epsilon_{\psi_{\Phi_{v}}} =0.1\). 
The above hyperparameter values were chosen to balance exploration with runtime and were held constant across all experiments.

While \ac{vlaps}'s search procedure could, in principle, be allowed to continue indefinitely, we limit the algorithm's maximum wall-clock execution time to \(600\) seconds per task to ensure that all experiments can be run in practice.
That is, if \ac{vlaps} has not successfully completed a task within \(600\) seconds, the attempt is considered a failure.
This timeout is not intended to reflect a practical deployment constraint, but rather to make large-scale evaluation tractable and to demonstrate the performance improvements that \ac{vlaps} can achieve even under finite compute budgets.

\subsubsection{Effect of Base VLA Quality on VLAPS Performance and Search Efficiency}
\begin{figure}
    \centering
    \includegraphics[]{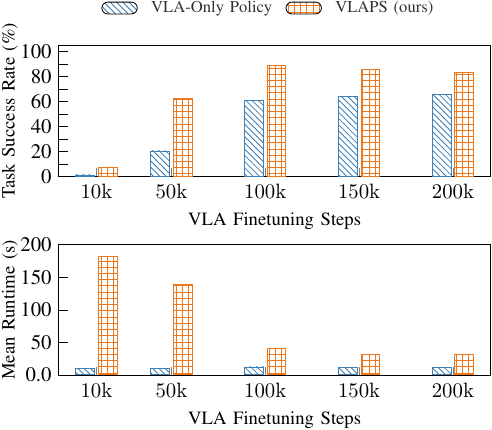}
    \caption{
        Performance of \ac{vlaps} and the \ac{vla}-only policy in Libero, as a function of the number of finetuning steps of the underlying \ac{vla} model.
        Top: Task success rate. 
        Bottom: Mean algorithm runtime to complete each task. 
        Runtimes are reported only for successful task evaluations. Failed \ac{vlaps} evaluations consistently reach the \(600\)s search timeout.
        For each \ac{vla} checkpoint, both \ac{vlaps} and the \ac{vla}-only policy are evaluated on \(1000\) total tasks, drawn from five Libero task suites: Libero-Spatial, Libero-Goal, Libero-Object, Libero-10, and Libero-90. 
        Each suite contributes 10 distinct tasks, each of which we test from ten different initial conditions.
    }
    \label{fig:octo_bar_chart}
\end{figure}

To examine the relationship between \ac{vlaps}' performance and that of the underlying \ac{vla} policy, we use the finetuning procedure as a controllable proxy for model quality.
That is, we save model checkpoints at \(10\textrm{k}, 50\textrm{k}, 100\textrm{k}, 150\textrm{k}\), and \(200\textrm{k}\) gradient steps during finetuning, and evaluate \ac{vlaps} using each checkpoint.
Figure \ref{fig:octo_bar_chart} illustrates \ac{vlaps}' resulting performance values, alongside the performance of the same models when deployed as standalone \ac{vla}-only policies.
More specifically, for each model checkpoint the figure shows the method's success rate over a total of \(1000\) tasks drawn from all five Libero task suites, as well as the average algorithm runtime for each successful task evaluation.
Table \ref{tab:libero-multicol} further elaborates on these results, detailing performance metrics of the evaluated checkpoints separately on each of the various Libero task suites. 
From these results, we observe that \ac{vlaps} consistently outperforms the \ac{vla}-only baseline, that these gains remain substantial even when the base \ac{vla} has a very low success rate, and that as the base model’s quality improves, \ac{vlaps}’s search procedure becomes markedly faster.
We also observe that \ac{vlaps} is capable of boosting the performance of small \ac{vla} policies to match that of much larger, state-of-the-art models.
We discuss these points in more detail in \S \ref{sec:experimental_results}.

\subsection{Experimental Results and Discussion}
\label{sec:experimental_results}

\begin{table}[t]
\caption{
    Task success rate and mean algorithm runtime across Libero task suites and \ac{vla} finetuning checkpoints.
    Each reported success rate is computed over \(100\) separate trials consisting of \(10\) distinct tasks and \(10\) different initial conditions per task.
    Runtimes are reported only for successful task evaluations. 
    Failed \ac{vlaps} evaluations consistently reach the \(600\)s search timeout.
}
\label{tab:libero-multicol}
\centering
\renewcommand{\arraystretch}{1.2}
\begin{tabular}{
    >{\centering\arraybackslash}p{1.4cm} 
    >{\centering\arraybackslash}p{1.0cm}   
    *{4}{>{\centering\arraybackslash}p{0.8cm}}  
}
\toprule
\textbf{Task Suite} & \textbf{ckpt.} 
& \multicolumn{2}{c}{\textbf{Success Rate (\%)}} 
& \multicolumn{2}{c}{\textbf{Mean Runtime (s)}} \\
\cmidrule(lr){3-4} \cmidrule(lr){5-6}
& 
& \textbf{VLA} & \textbf{VLAPS}
& \textbf{VLA} & \textbf{VLAPS} \\
\midrule

{Spatial}
& 10k    & 0  & 0  & N/A & N/A \\
& 50k    & 34 & 97 & 8.9 & 71.9 \\
& 100k   & 83 & 99 & 8.5  & 17.9 \\
& 150k   & 87 & 97 & 8.0  & 12.4 \\
& 200k   & 86 & 98 & 7.9  & 13.5 \\
\midrule

{Goal}
& 10k    & 3 & 24  & 6.8 & 136.3 \\
& 50k    & 50 & 86 & 6.7 & 54.3 \\
& 100k   & 87 & 94 & 6.5  & 19.8 \\
& 150k   & 87 & 95 & 6.6  & 18.9 \\
& 200k   & 91 & 91 & 6.6  & 15.3 \\
\midrule

{Object}
& 10k    & 0  & 2  & N/A & 289.0 \\
& 50k    & 6 & 73 & 9.0 & 144.8 \\
& 100k   & 32 & 82 & 7.9  & 41.8 \\
& 150k   & 22 & 63 & 8.4  & 38.8 \\
& 200k   & 19 & 54 & 9.1 & 39.8 \\
\midrule

{90}
& 10k    & 3  & 9  & 13.2 & 121.2 \\
& 50k    & 12 & 51 & 15.3 & 147.3 \\
& 100k   & 65 & 94 & 14.7 & 37.8 \\
& 150k   & 69 & 94 & 14.5  & 36.5 \\
& 200k   & 70 & 91 & 15.1  & 32.6 \\
\midrule

{10}
& 10k    & 0  & 0  & N/A & N/A \\
& 50k    & 0 & 6 & N/A & 273.8 \\
& 100k   & 37 & 74 & 23.3  & 87.5 \\
& 150k   & 56 & 81 & 22.1  & 47.8 \\
& 200k   & 63 & 84 & 20.6  & 58.3 \\


\bottomrule
\end{tabular}
\end{table}

\subsubsection{\ac{vlaps} Provides Strict Improvements Over \ac{vla}-Only Baseline Policies in All Experiments}
The bar chart on the left of Figure \ref{fig:octo_bar_chart} illustrates the consistent improvements to task success rate that \ac{vlaps} enjoys over the \ac{vla}-only baseline method.
These improvements are not only consistent---occurring for every tested checkpoint of the base model---but also substantial: \ac{vlaps} boosts overall task success rate by \(42\%\) for the \ac{vla} finetuned for \(50k\) steps, and by at least \(20\%\) for each of the \(100k\)-, \(150k\)-, and \(200k\)-step checkpoints.

Figure \ref{fig:filmstrips} illustrates several example tasks where \ac{vlaps} succeeds while the underlying \ac{vla} policy fails.
When examining such individual tasks,
we qualitatively observe that \ac{vlaps} more frequently avoids small errors---such as dropping an object or moving the robot into uncommon states not well represented in the demonstration data---that can derail an entire trial.

\subsubsection{\ac{vlaps} Yields the Largest Relative Gains when Augmenting \ac{vla}-Only Policies with Low Baseline Success Rates}
When examining the results of the \(50k\)-step model checkpoint in Table \ref{tab:libero-multicol}, we observe that \ac{vlaps} boosts the \ac{vla}'s success rate from \(6\%\) to \(73\%\) on the \textit{Libero-Object} task suite.
For this same model checkpoint, \ac{vlaps} also increases performance from \(34\%\) to \(97\%\) and from \(50\%\) to \(86\%\) on the Libero-Spatial and Libero-Goal task suites, respectively.
These results indicate that \ac{vlaps}' planning procedure can substantially increase the performance of less specialized models without additional finetuning, even when the quality of the initial policy is poor.
However, the \(10k\)-step checkpoint results highlight the limits of this effectiveness: when the base \ac{vla} policy fails entirely, \ac{vlaps} does not succeed within the allocated search time.

\subsubsection{\ac{vlaps} Automatically Allocates More Time to Search When the Base Policy Fails}
We observe from the bar chart on the right of Figure \ref{fig:octo_bar_chart} that as the performance of the underlying \ac{vla} policy improves, the average time \ac{vlaps} spends on search decreases sharply. 
Intuitively, this occurs because when the \ac{vla} policy alone can solve a task, \ac{vlaps} will return the \ac{vla}-suggested solution very quickly.
It is only when the base policy fails that \ac{vlaps} expands its search tree, effectively allocating additional compute to find an alternate, successful trajectory. 
We remark that the model checkpoints for which \ac{vlaps} spends the most time searching (on average) are exactly those for which base policy performs poorly---and where search yields the largest relative performance gains.

We note that while in this work we focus on finetuning as a controllable proxy for policy quality, we expect that the observed trends will generalize beyond this specific setting. 
For example, \ac{vlaps} is likely to benefit from improved task success rates and reduced search times as more capable \ac{vla} policies are developed---whether through alternative architectures, training datasets, or observation spaces---not only through finetuning on task-specific data.

\subsubsection{\ac{vlaps} Boosts Small Model Performance to that of Much Larger, State-of-the-Art Models}
Table \ref{tab:libero-compare} compares the task success rates of Octo, \ac{vlaps} (using Octo as the base \ac{vla} policy), and \(\pi_0\)-FAST \cite{pertsch2025fast}.
Both Octo and \(\pi_0\)-FAST were finetuned on demonstrations from the four Libero task suites that are included in the table.

In comparison with \(\pi_0\)-FAST---a state-of-the-art, 3.3 billion parameter \ac{vla} model finetuned from a PaliGemma VLM \cite{beyer2024paligemma} backbone---Octo is relatively small, containing only 93 million parameters.
While Octo alone achieves relatively low performance, \ac{vlaps} boosts this performance to a level that is comparable with the much larger \(\pi_0\)-FAST.
This highlights \ac{vlaps} as an effective method to achieve large-model performance from smaller, more compact \ac{vla}s.

\subsubsection{The Proposed \ac{vla}-Driven Action Sampling Enables Efficient Search That Would Otherwise Be Intractable}
Finally, we note that without the context-aware refinement from \S\ref{sec:defining_search_space}, the branching factor of search is \(b = |\macroActionLibrary| \approx 2000\).
Across our experiments, we observed typical search depths on the order of \(d=100\) before receiving a sparse reward for task completion. 
A full \(m\)-ary tree of height \(d\) has \(\frac{b^{d+1}-1}{b-1} \approx b^{d}\) nodes; with \(b=2000\) and \(d=100\) this would result in \(2000^{100}\) total nodes, which clearly renders direct search intractable. 
Meanwhile, uniform sampling would require on the order of \(b^{d}\) trials in expectation to sample a successful trajectory, which is far beyond any reasonable compute budget.
By contrast, \ac{vlaps} focuses search on trajectories that are similar to those initially suggested by the \ac{vla} policy, and consistently finishes search within the finite compute budgets described in \S \ref{sec:results_experimental_procedure}.

\begin{table}[t]
\caption{
    Task success rates (\%) on Libero task suites.
    With \ac{vlaps}, Octo (97M) achieves comparable performance to \(\pi_0\)-FAST (3.3B), a much larger state-of-the-art model.
    Without \ac{vlaps}, Octo performs relatively poorly by comparison.
}
\label{tab:libero-compare}
\centering
\renewcommand{\arraystretch}{1.2}
\begin{tabular}{
    >{\centering\arraybackslash}p{1.2cm}  
    >{\centering\arraybackslash}p{1.8cm}  
    >{\centering\arraybackslash}p{1.6cm}  
    >{\centering\arraybackslash}p{1.8cm}  
}
\toprule
\textbf{Task Suite} 
& \textbf{Octo (93M Params)} 
& \textbf{Octo + VLAPS (Ours)} 
& \textbf{\(\pi_0\)-FAST (3.3B Params)} \\
\midrule
Spatial   & 83 & 99 & 96 \\
Goal      & 87 & 94 & 96 \\
Object    & 32 & 82 & 99 \\
Libero-10 & 37 & 74 & 71 \\
\bottomrule
\end{tabular}
\end{table}

\section{Limitations and Future Work}
The proposed approach requires access to a world model to simulate sequences of future outcomes. 
This model may be a high-fidelity robot simulator, a learned dynamics model coupled with a visual rendering engine, or a video prediction model.
We expect that frequent feedback from observations could help mitigate any sim-to-real errors when \ac{vlaps} is used in a receding-horizon fashion. 
In this work, however, we focus on the core challenge of integrating model-based search with pre-trained \ac{vla} policies, assuming an accurate model is available. 
Developing methods to jointly learn a world model and use it for planning—similar to \textit{MuZero} \cite{schrittwieser2020mastering}—as well as studying the impact of model-environment mismatch, are important directions for future work.

VLAPS improves the performance of pre-trained \ac{vla} policies without requiring additional training. 
However, it incurs additional calls to the \ac{vla} model at test time during tree search, and the long inference times of current \ac{vla} models can thus lead to increased planning latency. 
However, optimizations to the algorithm's implementation can help to mitigate these challenges in practice.
For instance, \ac{vla} queries can be queued and processed in batches, while the expansion and rollout phases of the search algorithm can also be executed in parallel. 
Additionally, the branching factor and maximum tree depth are configurable parameters that provide direct control over test-time computational cost. 
Looking forward, \ac{vla} models themselves are likely to benefit from significant future improvements to efficiency. 
Techniques such as model quantization and distillation, for example, may substantially reduce inference times, making VLAPS increasingly practical for operational deployment.

\section{Conclusions}
\label{sec:conclusions}
We present \acf{vlaps}---a framework that integrates model-based search with pre-trained \acf{vla} models to enable generalist robot policies that are robust, and capable of reasoning over future action outcomes.
By using a \ac{vla} model to define context-aware priors over the search space, \ac{vlaps} enables efficient search over robot actions, even in environments with large action spaces and sparse rewards.
Our experiments in language-specified robot reasoning and manipulation tasks demonstrate that \ac{vlaps} consistently and significantly outperforms \ac{vla}-only baseline policies.
Importantly, \ac{vlaps} requires no additional training and is agnostic to the specific \ac{vla} model used---our results indicate that its performance will improve alongside future advances in pre-trained \ac{vla} models.
These results highlight VLAPS as a promising direction for increasing test-time compute and model-aware planning capabilities in \ac{vla} policies.

\section*{Acknowledgment}
We acknowledge funding support from the Natural Sciences and Engineering Research Council (NSERC) of Canada and the Canadian Institute for Advanced Research (CIFAR), as well as compute support from the Digital Research Alliance of Canada, Mila IDT, and NVIDIA.

\printbibliography

\end{document}